\pdfoutput=1

\documentclass[11pt,a4paper]{article}
\usepackage[nohyperref]{naacl2021}
\usepackage{hyperref}
  \usepackage{adjustbox}
\usepackage{subcaption}
\usepackage{booktabs}
\usepackage{times}
\usepackage{latexsym}
\usepackage[T1]{fontenc}
\usepackage{tabularx}
\usepackage{placeins}

\usepackage{stfloats}

\usepackage{microtype}

\usepackage{amssymb}
\usepackage{wasysym}
\usepackage{tabulary}

\def\Snospace~{\S{}}
\def\Ssmallspace~{\S{}\,}

\definecolor{green}{RGB}{44, 160, 44}
\definecolor{orange}{RGB}{255, 127, 14}
\definecolor{blue}{RGB}{31, 119, 180}
\definecolor{purple}{RGB}{148, 103, 189}

 \title{Measuring the `I don't know' Problem\\ through the Lens of Gricean Quantity}

\author{Huda Khayrallah  \\
  Johns Hopkins University \\
  \texttt{huda@jhu.edu} \\\And
  Jo\~ao Sedoc \\
   New York University \\
  \texttt{jsedoc@stern.nyu.edu} \\}

\date{}

\begin{document}
\maketitle

\begin{table*}[hb]
\centering
\begin{tabularx}{\textwidth}{p{0.12\linewidth} | p{0.15\linewidth}| p{0.33\linewidth} | p{0.3\linewidth} }
\toprule

Maxim & Definition & Violated by... & Prompt: What color is grass?\\ \midrule
\textsc{Quantity} & Be informative.               &  not answering a question (fully), \newline or giving too much information.          & I don't know.                                                                                                                                                           \\ \midrule
\textsc{Quality}  & Be truthful.                  &  lying, or saying something \newline \mbox{without} evidence.                & Grass is purple.                                                                                                                                                        \\ \midrule
\textsc{Relation} & Be relevant.                  &  off-topic responses.                                        & I like pizza.                                                                                                                                                           \\ \midrule
\textsc{Manner}   & Be clear, brief, \newline and orderly. & disfluent responses & is green grass usually.

\\
\bottomrule
\end{tabularx}
\caption{Gricean maxims, with examples of how they can be violated for the prompt `What color is grass?'}
\label{tab:grice_ex}
\end{table*}

\begin{abstract}
We consider the intrinsic evaluation of neural generative dialog models  through the lens of Grice's Maxims of Conversation~(\citeyear{LogicandConversation}).
Based on the maxim of Quantity (be informative), we propose 
Relative Utterance Quantity (RUQ) to diagnose the `I don't know' problem, in which a dialog system produces generic responses. The linguistically motivated RUQ diagnostic compares the model score of a generic response to that of the reference response. We find that for reasonable baseline models, `I don't know' is preferred over the reference the majority of the time, but this can be reduced to less than 5\% with hyperparameter tuning. RUQ allows for the direct analysis of the `I don't know' problem, which has been \emph{addressed} but not \emph{analyzed} by prior work.
\end{abstract}

\section{Introduction}
Neural generative dialog models have a tendency to produce generic, safe responses, such as `I don't know' \cite{serban-etal-2016-building,li-etal-2016-diversity}. 
The repetition of such phrases is annoying to users, and contributes nothing to the conversation.

Evaluating chatbots is an active area of research, partly due to their open-ended nature 
\cite{hashimoto-etal-2019-unifying,sedoc-etal-2019-chateval,li2019acuteeval,mehri-eskenazi-2020-usr,deriu-etal-2020-spot}. 
To the best of our knowledge, 
no prior work focuses on \emph{analyzing} systems for generic, safe responses, such as `I don't know.' While prior work 
\cite{li-etal-2016-diversity,li-etal-2016-deep,csaky-etal-2019-improving,welleck-unlikelihood} \emph{addresses} the  `I don't know' problem, the lack of \emph{analysis} leaves it unclear if a method improves models by mitigating \emph{this} problem, or another.

One linguistic framework for analyzing  conversations is \citeauthor{LogicandConversation}'s Cooperative Principle (\citeyear{LogicandConversation}), which consists of
 Maxims of Conversation that function as guidelines for effective communication. \citeauthor{LogicandConversation} considered conversations between humans, but there has also been some exploration in  NLP  \cite{Bernsen1996Cooperativity,harabagiu1996testing,qwaider-etal-2017-trentoteam,jwalapuram-2017-evaluating}. 

We discuss each of the categories of maxims and the ways a chatbot might violate them in \autoref{tab:grice_ex}.

We propose a novel automatic diagnostic inspired by the Gricean \textsc{Quantity} maxim. Relative Utterance Quantity checks if the model favors a generic response (such as `I don't know.') over the reference it was trained on for each prompt. We apply our diagnostic to a method designed to address this problem \cite{csaky-etal-2019-improving}, and find that method does mitigate it, though not by as much as a hyperparameter search.

\FloatBarrier
 
 \section{Relative Utterance Quantity (RUQ)}

If a system responds  `I don't know.' when it could have given a better or more informative answer, this is by definition a violation of \textsc{Quantity}.  
Based on this interpretation 
we propose a method for diagnosing the problem. We compare the model score of producing `I don't know.' to the  model score of producing the  reference response. This can be done on the training data, or the test data. Particularly on the training data, we should expect the model to `know' 
the data it was trained on and therefore score it higher than `I don't know.' 

We propose two diagnostic measures to compute the Relative Utterance Quantity of a model: 
(1)~We plot the average model score for each token across sentences. We compare the original reference, beam search  output, and two `I don't know' (IDK) variants: `I don't know.' and `I don't know what to do.' allowing for the visualization of the relative gap in scores at different points in the sentence. (2)~We compute the (length normalized) model score for `I don't know.' and the reference of each training prompt, and count how many times the reference is preferred. We denote the later as RUQ score.
Both generalize to other generic responses, as might be appropriate for other corpora or other languages.

If there are multiple references we would recommend comparing the lowest likelihood   reference for RUQ score, since all valid references should be better than I don’t know.

We note that RUQ captures some types of \mbox{\textsc{quantity}} violations, but not all violations of this maxim.  

\section{Data}
Following \citet{khayrallah-sedoc-2020-SMRT}, we train and evaluate on DailyDialog \cite{li-etal-2017-dailydialog},\footnote{ As released by \href{https://github.com/facebookresearch/ParlAI/tree/1e905fec8ef4876a07305f19c3bbae633e8b33af}{ParlAI} \cite{miller-etal-2017-parlai}. The ParlAI release of DailyDialog is tokenized and lowercased. Following \citet{khayrallah-sedoc-2020-SMRT} we detokenize and recase the DailyDialog data for training.}  which consists of $\sim\,$80,000 turns of English-learners practicing `daily dialogues' in various contexts, e.g., chatting about vacation or food.

We also use Entropy-Based Data Filtering \cite{csaky-etal-2019-improving}, which filters out high entropy utterances\footnote{Prompts that solicit many different responses and responses that can apply to many different prompts.} with the goal of removing generic ones. We use the  recommended filtering threshold of 1.0 and `IDENTITY' clustering. We filter based on their `source', `target', and `both' settings. We consider `target' as the baseline, as they find it works best. 
We denote models trained on DailyDialog as \textsc{dd} and models trained on \citeauthor{csaky-etal-2019-improving}'s entropy filtered version as \textsc{ef}.

\section{Evaluation Metrics}
\label{sec:eval}
\subsection{Standard Automatic Metrics}

\label{sec:auto_eval}
We use the single-reference and multi-reference\footnote{For RUQ, we only use the original single-reference.} automatic evaluation framework for DailyDialog released by \citet{gupta-etal-2019-investigating},\footnote{\href{https://github.com/prakharguptaz/multirefeval/tree/384d39f80c94448fffd450c9a6fe91903db3f325}{github.com/prakharguptaz/multirefeval}} which is computed using \textsc{nlg-eval} \cite{sharma2017nlgeval}.\footnote{\href{http://github.com/Maluuba/nlg-eval}{github.com/Maluuba/nlg-eval}}
We primarily consider multi-reference METEOR \cite{lavie-agarwal-2007-meteor}; 
see Appendix~\ref{app:full} for all metrics.\footnote{For reading ease, we report metrics scaled between 0 and 100 rather than 0 and 1.}

\subsection{Human Evaluation}
For human evaluation of the different systems we use crowdworkers on Amazon Mechanical Turk to judge the  fluency, coherence, and interestingness of utterances on a 1-5 Likert scale (see Appendix~\ref{sec:app-hit} for full details) for 100 randomly sampled evaluation set prompts. Four annotators judge the responses from all systems for each prompt in a single turn context.  We remove any annotators with a linear Cohen's Kappa $<$ 0.1 from the results.

\section{Models}
\label{sec:params}
Following \citet{khayrallah-sedoc-2020-SMRT}, 
we train Transformer \cite{transformer} chatbots in \textsc{fairseq} using parameters from the \textsc{flores} benchmark for low-resource MT \cite{guzman-etal-2019-flores}:\footnote{See \autoref{app:dialog_models} for full details for replication.}
$5$-layer encoder and decoder, $512$ dimensional embeddings, and $2$ encoder and decoder attention heads. The default regularization parameters are $0.2$ label smoothing \cite{labelsmooth}, $0.4$ dropout, and 0.2 	attention \& ReLU dropout.

\begin{figure*}[ht]
     \centering
     \begin{subfigure}[b]{0.267\textwidth}
 \includegraphics[width=\textwidth]{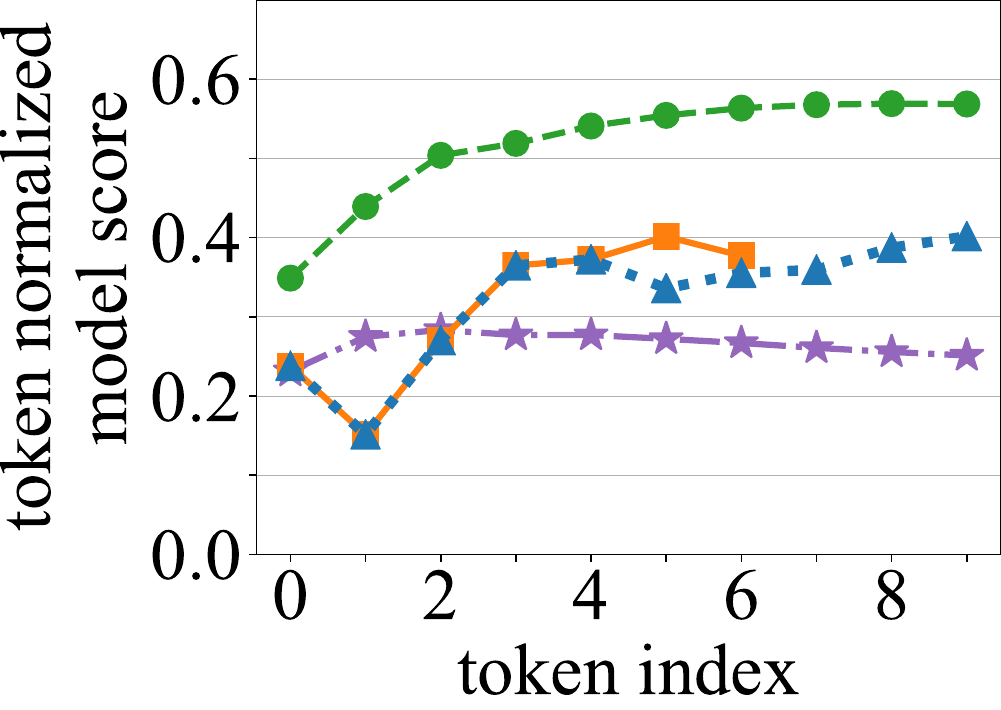}
      \caption{\textsc{dd-base} train \textsc{ruq}}\vspace{10pt}
     \label{fig:flores_train}
     \end{subfigure}
      \hfill
 \begin{subfigure}[b]{0.23\textwidth}
 \includegraphics[width=\textwidth]{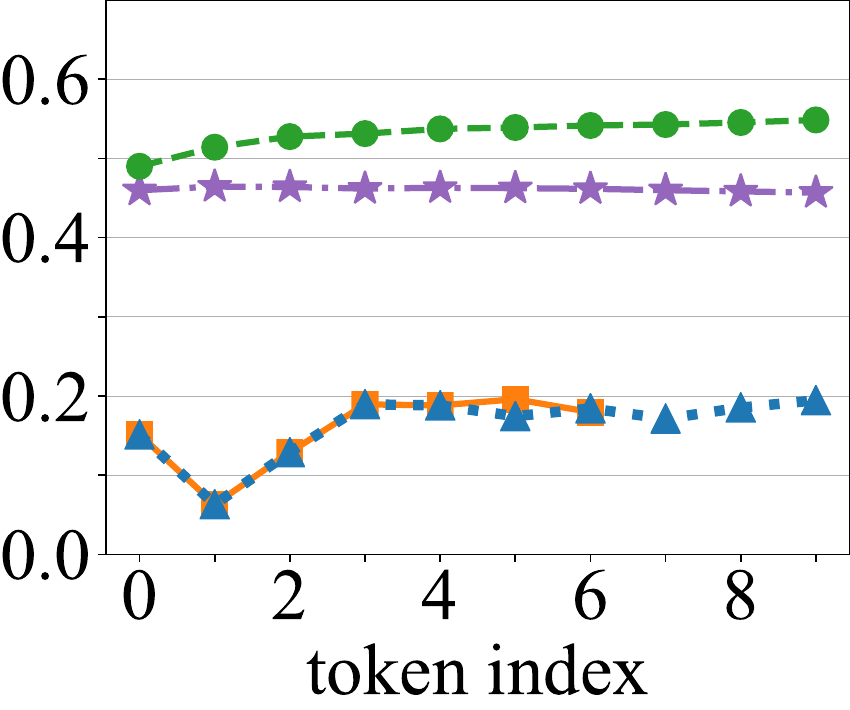}
 \caption{\textsc{dd-best} train \textsc{ruq}}\vspace{10pt}
     \label{fig:floresbest_train}
     \end{subfigure}
      \hfill
    \begin{subfigure}[b]{0.23\textwidth}
 \includegraphics[width=\textwidth]{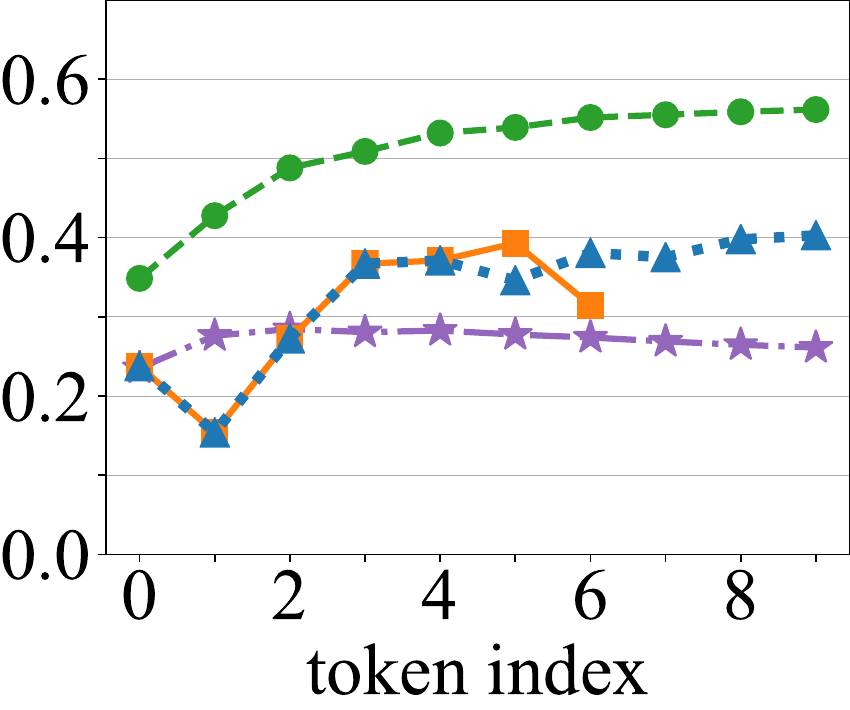}
\caption{\textsc{ef-base} train \textsc{ruq}}\vspace{10pt}
     \label{fig:floresfilter_train}
     \end{subfigure}
      \hfill
 \begin{subfigure}[b]{0.23\textwidth}
 \includegraphics[width=\textwidth]{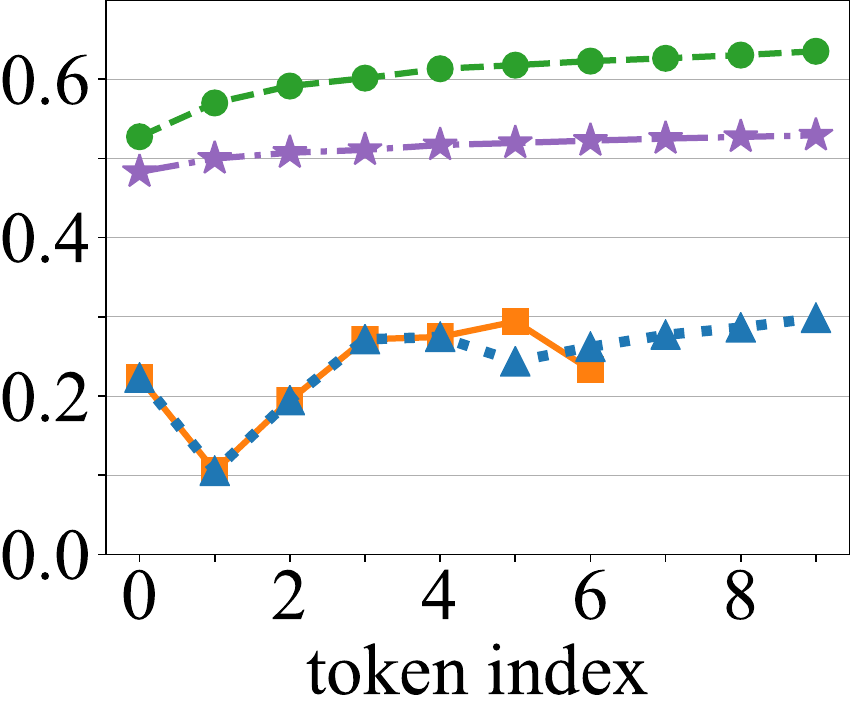}
  \caption{\textsc{ef-best} train \textsc{ruq}} \vspace{10pt}
          \label{fig:floresfilterbest_train}
     \end{subfigure}
     
         \centering
     \begin{subfigure}[b]{0.267\textwidth}
 \includegraphics[width=\textwidth]{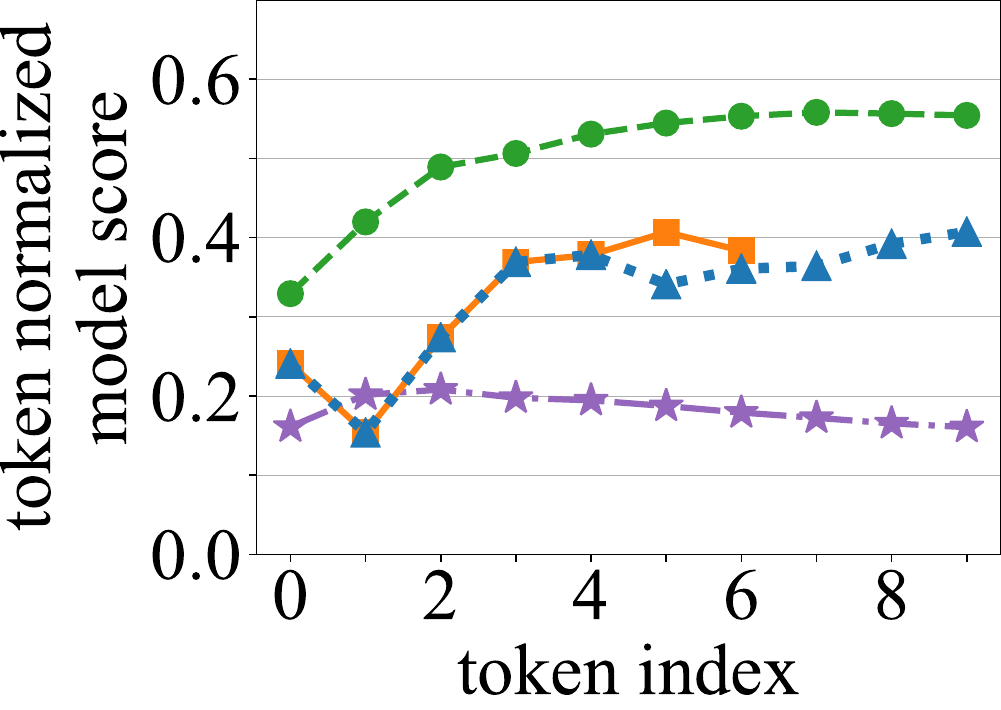}
  \caption{\textsc{dd-base} test \textsc{ruq}}
     \label{fig:flores_test}
     \end{subfigure}
      \hfill
 \begin{subfigure}[b]{0.23\textwidth}
 \includegraphics[width=\textwidth]{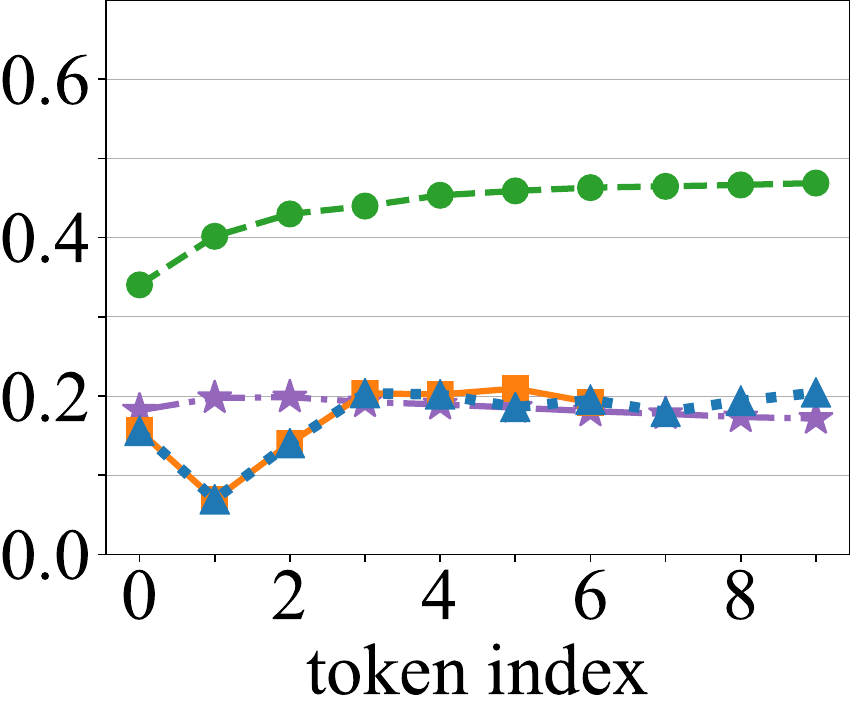}
  \caption{\textsc{dd-best} test \textsc{ruq}}
     \label{fig:floresbest_test}
     \end{subfigure}
      \hfill
    \begin{subfigure}[b]{0.23\textwidth}
 \includegraphics[width=\textwidth]{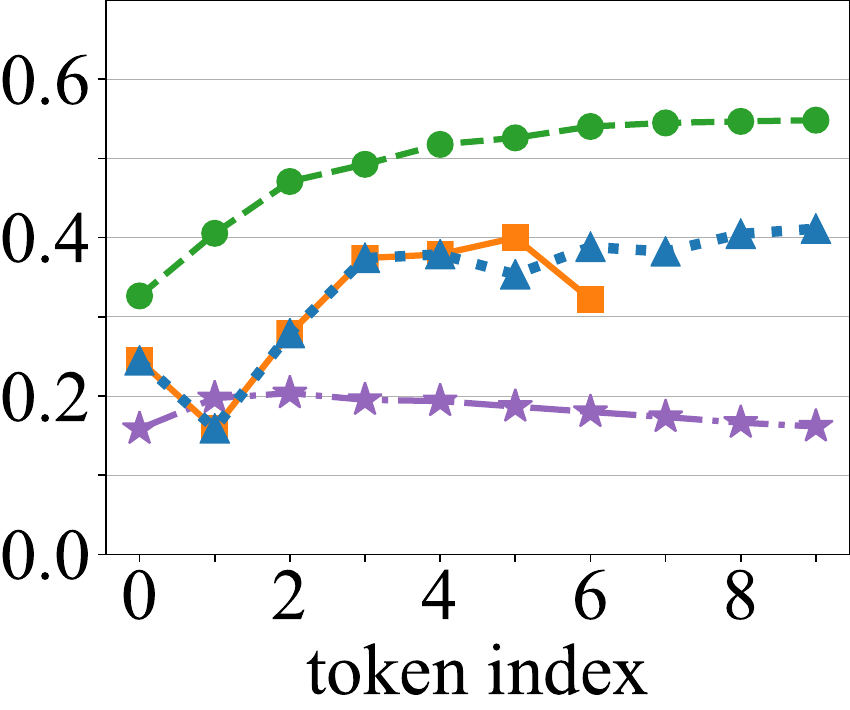}
   \caption{\textsc{ef-base} test \textsc{ruq}}
     \label{fig:floresfilter_test}
     \end{subfigure}
      \hfill
 \begin{subfigure}[b]{0.23\textwidth}
 \includegraphics[width=\textwidth]{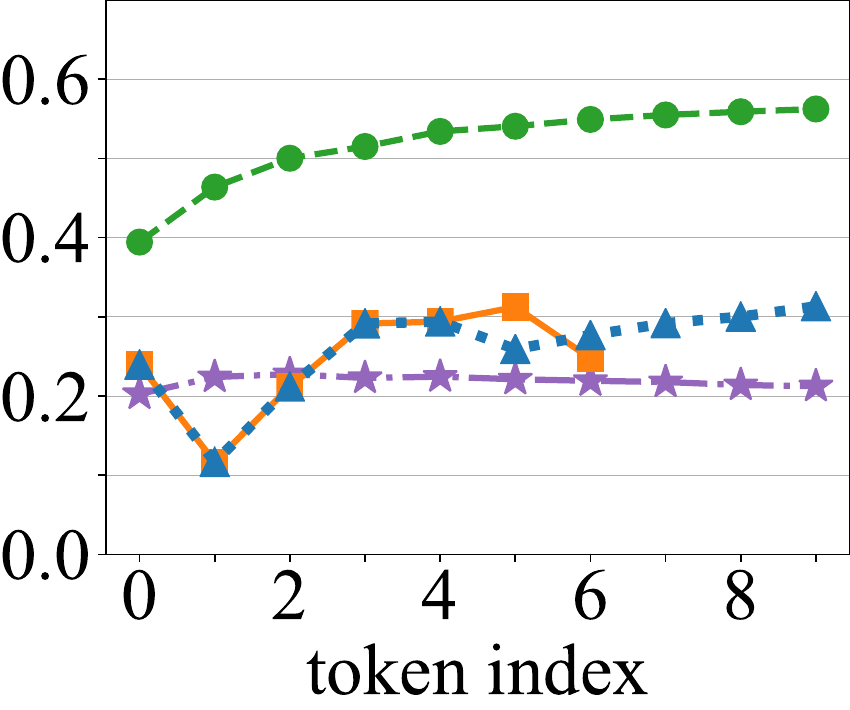}
    \caption{\textsc{ef-best} test \textsc{ruq}}
     \label{fig:floresfilterbest_test}
     \end{subfigure}
         \caption{RUQ plots on the train (top) and test (bottom) data. We plot the token normalized model score for the \textcolor{purple}{reference ($\bigstar$)}, the \textcolor{green}{beam-search output (\CIRCLE)}, \textcolor{orange}{`I don't know.' ($\blacksquare$)}, and \textcolor{blue}{`I don't know what to do.' ($\blacktriangle $)}. Points are per (subword) token, and averaged over all prompts.}
        \label{fig:RUQ}

\end{figure*}

\subsection{Hyperparameter Sweep}
Some kinds of regularization (e.g., label smoothing and subword vocabularies) are not universally used in dialog.\footnote{For example popular toolkits for dialog (e.g., Hugging Face \cite{wolf-etal-2020-transformers} and ParlAI \cite{miller-etal-2017-parlai}) do not implement label smoothing.}
Since we are concerned with the model over-fitting on IDK,
 we perform a hyperparameter sweep of regularization parameters, including  SentencePiece  \cite{kudo-richardson-2018-sentencepiece} vocabulary size, learning rate, dropout,	attention \& relu dropout, and  label smoothing.\footnote{See Appendix~\ref{app:sweep} for more hyperparameter details.}

We denote models trained with the \textsc{flores} hyperparameters as \textsc{base}, and the best model from the hyperparameter searches for each data type (as selected by multiple-reference METEOR) as \textsc{best}.

\begin{table}[]
\centering
\begin{tabular}{l|cc}
\toprule
 training data & \textsc{base} & \textsc{best} \\
\midrule
 \textsc{DailyDialog}  & 12.7 &17.8  \\
  \textsc{Entropy-Filtered}& 13.2 & 17.2 \\
\bottomrule
\end{tabular}

\caption{Multi-reference METEOR for the four systems we analyze in this work.  \textsc{best} models are the result of the hyper parameter sweeps.}

\label{tab:meteor}
\end{table}

We report the multi-reference METEOR scores for the \textsc{base} and \textsc{best} sysems in \autoref{tab:meteor}.\footnote{We report hyperparameters of these models and their performance on the full set of automatic metrics in \autoref{app:full}.}
For the DailyDialog data we find that hyperparameter tuning can improve multiple-reference METEOR from 12.7 (\textsc{dd-base}) to 17.8 (\textsc{dd-best}). 

We perform the same hyperparameter sweep after performing entropy filtering \cite{csaky-etal-2019-improving} on the data, but we find that the best model is still \textsc{dd-best}. 
Without hyperparameter tuning, entropy filtering improves performance by $\sim$0.5 on multi-reference METEOR, but the improvement by hyperparameter sweeping is much larger (5.1 points).\footnote{We note that \citet{csaky-etal-2019-improving}---who proposed entropy filtering and an observed a 1 BLEU point improvement from using it (we observed a 0.3 improvement in single reference BLEU)---did not use any subwords units; they  used a total vocab size of 16k. Our 10 best systems all had Sentencepiece vocab sizes of 2k, 4k, or 8k, so perhaps this difference may explain the discrepancy between their results and our replication. We note that for the 3 metrics which we believe our evaluations are comparable---single reference Embedding Average Cosine Similarity, and single reference	Vector Extrema Cosine Similarity---our baseline outperforms their results. The BLEU scores are not directly comparable because they report sentence BLEU, while we report corpus BLEU following \citet{gupta-etal-2019-investigating}.}

We did a very thorough sweep (including values we expected to perform poorly), which led to some general takeaways:
Using a subword vocabulary (of 4-8k) is helpful. (2) Label smoothing interacts with subword vocabulary size, but is also helpful.

\section{Relative Utterance Quantity}
\subsection{RUQ Plots}

We show plots for the four models in \autoref{fig:RUQ}.  
We plot the  token normalized model score for reference and `I don't know.' For additional comparison, we also plot the model scores for the beam-search output and `I don't know what to do.'
Overall, we observe that for the \textsc{base} models the IDKs are higher probability than the reference, even on the training data. This is problematic, because the model is ranking a response that is not providing enough \textsc{Quantity} of information higher than the reference despite the fact that it should `\emph{know}' the training data.
The relative difference in probabilities is much better in \textsc{dd-best} than \textsc{dd-base}, particularly on the training set. 
Simply entropy filtering the data alone does not fix the problem.

\subsection{RUQ scores}
We summarize \textsc{Quantity} in a single statistic by counting how many times the reference has a higher probability than `I don't know.' on the training data.

Entropy filtering improves how often the reference is preferred to `I don't know.', but not by as much as the hyperparameter sweep does, see \autoref{tab:ruq-train} for the RUQ scores on the training data.\footnote{RUQ scores on the on the test data are reported in \autoref{app:full}.
The overall trend is same, but the absolute values lower.} For both \textsc{dd-base} and  \textsc{ef-base}, IDK is preferred over the reference response the model was trained on over half of the time (71.5\% for \textsc{dd}, 62.1\% for \textsc{ef}). 

\begin{table}[]
\centering
\begin{tabular}{l|cc}
\toprule
training data &\textsc{base}  & \textsc{best} \\
\midrule
\textsc{DailyDialog}   & 28.5\% & 95.3\% \\
\textsc{Entropy-Filtered}& 37.9\%  &89.2\% \\
\bottomrule
\end{tabular}
\caption{Training data RUQ scores. Entropy filtering improves how often the reference is preferred to `I don't know.', but by less than the hyperparameter sweeps (which are denoted \textsc{best}).}    
\label{tab:ruq-train}
\end{table}

\begin{table}[]
\centering

\begin{adjustbox}{max width=\linewidth}

  \addtolength{\tabcolsep}{-2pt}

\begin{tabular}{l|ccc}
 \toprule
& Fluency & Coherence & Interestingness\\
 
\midrule
Human                                    & 4.9     & 4.6   & 4.0      \\
\midrule
\textsc{dd-base}                                    & \textbf{4.8}     & 3.5   & 2.6     \\
     \textsc{dd-best}                              & \textbf{4.8}    & \textbf{3.8} & 2.7  \\\midrule
 \textsc{ef-base}                & 4.4     & 3.3    & 2.8     \\
 \textsc{ef-best}             & 4.4     & 3.1    &\textbf{3.3}     \\
\bottomrule
\end{tabular}
\end{adjustbox}
\caption{Average human judgement ratings on 1-5 pointwise scale for DailyDialog (\textsc{dd}) and the entropy filtered (\textsc{ef}) data. The result of the hyperparameter sweep is denoted \textsc{best}.
}
\label{tab:pointwise}
\end{table}

\subsection{Human Evaluation}
\autoref{tab:pointwise} shows human judgments of fluency,  coherence, and interestingness.\footnote{\autoref{sec:h2h} discusses head to head judgments. Models trained on the DailyDialog data are preferred over the filtered models, but there is no clear preference between base and best models.} 
The models trained on DailyDialog have higher fluency and coherence, while the models trained on the filtered data have higher interestingness. For both kinds of data, the hyperparameter tuning (as selected by METEOR) improved interestingness. Fluency did not change. Coherence was reduced for the filtered models and improved for the base model. Improved RUQ may be reflected in either interestingness or coherence, but other factors can influence those judgments. Therefore, measuring RUQ directly is important to measuring progress on the IDK problem.
\section{Discussion}
The relative RUQ rankings of the four systems we consider in this work are the same as the relative rankings by multi-reference METEOR, and \textsc{dd-best} (the single best model according to mulit-reference METEOR) is also the one with the highest RUQ score. 
Among all models in the hyperparameter sweep, RUQ is correlated with METEOR with Spearman's $\rho$ of 0.9 but this drops to 0.6 when considering only the top 20 systems, demonstrating that RUQ and METEOR do not capture the same phenomenon.
We note that RUQ on the training data does not require a particular (multi-reference) test set like most automatic evaluation metrics. RUQ simply diagnoses how well the model learned the training data compared to a generic response.

The model's relative preference of IDK over the (presumably) better reference response is not only a \textsc{Quantity} violation, but is also indicative of a fundamental problem with the models themselves, and should be fixed before decoding time (either by correcting the data, or by correcting the model).

\citet{csaky-etal-2019-improving} argue that the IDK problem is due to the one-to-many/many-to-one nature of dialog training data---if a single response applies to many different responses, it will become the canonical response. Therefore their entropy filtering method removes one-to-many/many-to-one pairs, by removing high entropy responses. While this data filtering reduces the problem, we found that the baseline model trained on the entropy filtered data (\textsc{ef-base}) still preferred IDK over the reference the majority of the time, suggesting opportunities for future research on the IDK problem.

\section{Related Work}

\paragraph{Gricean Maxims in NLP}
Gricean maxims have previously been discussed in NLP. \citet{Bernsen1996Cooperativity} examine the relationship between a new set of maxims for human-bot dialogs and relate them to Gricean maxims. They point out that these do not entirely overlap; however, the maxim of Quantity is preserved since unambiguous contributing responses are required in conversations in general. \cite{harabagiu1996testing} attempt to explicitly create an evaluation methodology using sets of primitive rules and WordNet. Our approach is different as RUQ is a diagnostic metric.

\citet{jwalapuram-2017-evaluating} propose a Gricean  dialog evaluation where humans rate performance on a Likert scale for each category.   
\citet {qwaider-etal-2017-trentoteam} consider the  \textsc{quantity}, \textsc{relation}, and \textsc{manner} maxims  for ranking community question answers. They use other NLP tools to evaluate if the response has key elements or named entities (\textsc{quantity/relation}), has high semantic similarity (\textsc{relation}), and includes/excludes positive/negative polarity terms (\textsc{manner}). 

\paragraph{Chatbot evaluation}
Automatic evaluations for dialog typically measure lexical or semantic similarity between a produced response and a reference, under the assumption that the reference is a good response and responses similar to it will be good as well. Since there are often multiple valid responses to a prompt, this can be extended to multiple references too. In contrast, in our work we compare a model's score of a reference to a model's score of a generic response for directed analysis. 

HUSE \cite{hashimoto-etal-2019-unifying} uses the model score combined with human judgments to evaluate diversity and quality, classifying a response as human- or machine-generated. Our work does not require human judgments, and  compares the model score of a generic response to the reference response. 

\citet{mehri-eskenazi-2020-unsupervised} also use scoring from a model.  
Whereas that work is using an external model, we propose an intrinsic diagnostic for a particular phenomenon.
Each serves a different purpose, and an advantage of our method is
our analysis does not require an external model, which might not be available in all languages and for all types of text.

\paragraph{Mitigating the IDK Problem}
A variety of approaches have been proposed to \emph{mitigate} the IDK problem. These include active post-processing methods such as MMI \cite{li-etal-2016-diversity}, as well as training data filtration  \cite{csaky-etal-2019-improving}, reinforcement learning \cite{li-etal-2016-deep} and unlikelihood training \cite{welleck-unlikelihood}. 
In our work, we propose an intrinsic model diagnostic to \emph{analyze} the problem.

\paragraph{MMI}
Maximum Mutual Information was proposed as a `Diversity-Promoting Objective Function' for dialog \cite{li-etal-2016-diversity}. 
MMI-bidi encourages the prompt to be predictable from the response, by using a reverse direction model. 
We argue this was not diversity broadly speaking, but actually tackling a \textsc{relevancy} problem, since it is scoring how predictable the prompt is from the response.

\citeauthor{li-etal-2016-diversity} demonstrate MMI improves performance, though recent work found that it does not always \cite{khayrallah-sedoc-2020-SMRT}.

\paragraph{Copying in Machine Translation}
\citet{pmlr-v80-ott18a} found that copying was overrepresented in the output of RNN NMT.
Using an analysis that inspired RUQ plots they compare the score of the beamsearch output to that of the copied source. They also consider the probability at each position in the output, and find the model is unlikely to start copying; however, after starting to copy continuing to copy has high probability.
We find IDK has a relatively high score from the start, though for some models the gap widens towards the end of the sentence.

\section{Conclusion}
  We reframe the IDK problem as a violation of the  Gricean maxim of \textsc{Quantity}, and introduce a new measure---Relative Utterance Quantity (RUQ)---which allows researchers to diagnose if their model is violating this particular conversational principle, and analyze methods that aim to address it.

We aim to encourage further discussion and research drawing on linguistic principles about 
 discourse and pragmatics for analysis of dialog models.

\pagebreak
\section*{Acknowledgments}
We thank Patrick Xia, Nathaniel Weir, Rachel Rudinger, and Claire Daniele 
 for their helpful comments and feedback on the paper. We additionally thank the reviewers for their insightful comments. 
 
This work was supported in part by DARPA KAIROS (FA8750-19-2-0034). The views and conclusions contained in this work are those of the authors and should not be interpreted as representing official policies or endorsements of DARPA or the U.S. Government.

\bibliography{anthology,eacl2021}
\bibliographystyle{acl_natbib}
\clearpage
\appendix

\section{Appendix}
\subsection{Hyperparameter Search}
\label{app:sweep}
We sweep SentencePiece  \cite{kudo-richardson-2018-sentencepiece} vocabulary size (1k,4k,8k,16k), learning rate (1e-2, 1e-3, 1e-4), dropout (0.0, 0.1, 0.2, 0.3, 0.4, 0.5, 0.6),	attention \& ReLU dropout (0.0, 0.1, 0.2, 0.3, 0.4), and  label smoothing (0.0, 0.1, 0.2, 0.3, 0.4, 0.5, 0.6, 0.7, 0.8).

\subsection{Standard Automatic Metrics}
\label{app:auto_eval}
In \autoref{app:full} we report the full automatic evaluation results of the 14 metrics across both the single reference and multi-reference evaluation from the the multi-reference automatic evaluation framework for DailyDialog released by \citet{gupta-etal-2019-investigating},\footnote{\href{https://github.com/prakharguptaz/multirefeval/tree/384d39f80c94448fffd450c9a6fe91903db3f325}{github.com/prakharguptaz/multirefeval}} which is computed using \textsc{nlg-eval}\footnote{\href{https://github.com/Maluuba/nlg-eval/tree/846166566bf0fdccbaa9e5b41da97147470b525b}{github.com/Maluuba/nlg-eval}} \cite{sharma2017nlgeval}. 
This includes word-overlap metrics: BLEU \cite{papineni-etal-2002-bleu}, METEOR \cite{lavie-agarwal-2007-meteor}, and ROUGE-L \cite{lin-2004-rouge} as well as embedding based metrics: SkipThought \cite{NIPS2015_5950}, embedding average \cite{forgues2014bootstrapping}, vector extrema, and Greedy Matching \cite{rus-lintean-2012-comparison}.
For reading ease, we report metrics scaled between 0 and 100 rather than 0 and 1. 

\subsection{Lexical Diversity}

The Gricean maxims focus on ensuring cooperation between speakers, but there is more to a conversation than cooperation---especially in an open ended conversation that might be had with a chatbot. This is where additional desiderata may come in to play, such as interestingness. One (indirect) automatic way of measuring interestingness is lexical diversity \cite{halliday1989spoken,Laufer1995}, by computing the n-gram type/token ratio \cite{li-etal-2016-diversity}.
We use the same spaCy\footnote{\url{spacy.io}} tokenization used in the automatic evaluation scripts (\autoref{app:auto_eval}).\footnote{\href{https://github.com/Maluuba/nlg-eval/tree/846166566bf0fdccbaa9e5b41da97147470b525b}{github.com/Maluuba/nlg-eval}}

\subsection{Pointwise Human Evaluation}
\label{sec:app-hit}
\begin{figure*}
    \centering
    \includegraphics[width=\textwidth]{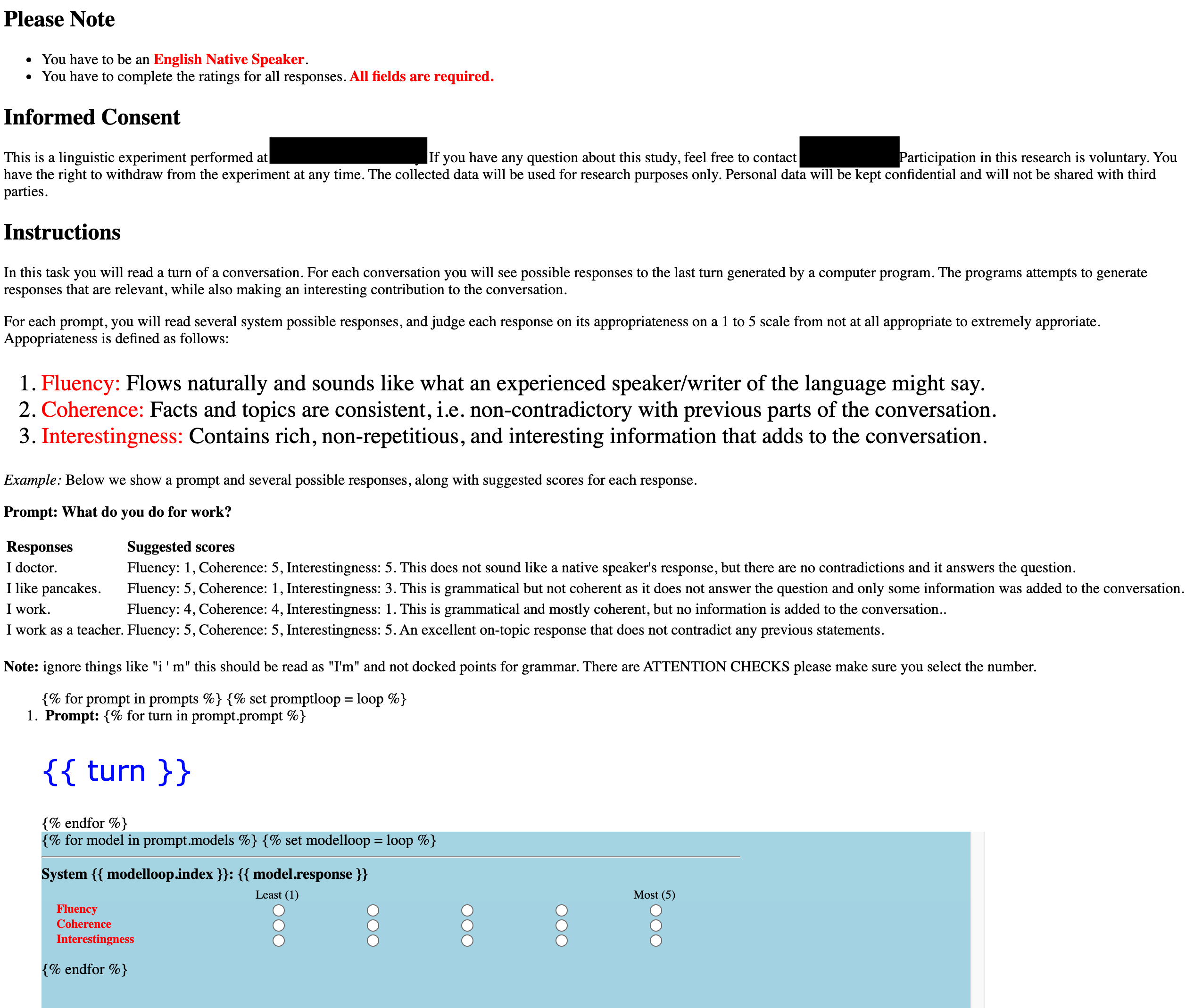}
    \caption{Instructions for AMT task.}
    \label{fig:amt}
\end{figure*}

We presented Amazon Mechanical Turk workers the task with 4 prompts and all system responses along with a human reference. The annotators had a maximum time allotted of 20 minutes. Our criteria for inclusion were over 500 approved HITs, an approval rate over 98\%, and location set to US. Each HIT was paid \$0.15 with an overlap of 4 annotators per HIT.\footnote{We  aimed to compensate the crowdworkers fairly (\$ 8 per hour) and did this by annotating a set of data ourselves to estimate the timing of the task} A screenshot of the HIT is in \autoref{fig:amt}.

\subsection{Head to Head Human Evaluation}
\label{sec:h2h}

In addition to the point-wise evaluation, we also test head-to-head pairwise performance on the evaluation set of 480 unique prompt/response pairs, as shown in \autoref{sec:h2h}. Models trained on the DailyDialog data outperform the filtered models, but there is no clear preference between base and best models.

\begin{table*}[]
\centering
\begin{tabular}{ll|ccc}
\toprule
M1&M2&M1&M2&tie\\
\midrule
\textsc{ef-best} & \textsc{dd-best}       & 38.5\% & 42.4\% & 19.1\% \\
\textsc{ef-base} & \textsc{ef-best} & 34.4\% & 34.0\% & 31.7\% \\
 \textsc{ef-base} & \textsc{dd-base}  &  37.0\% &41.6\% & 21.4\% \\
\textsc{dd-base}       & \textsc{dd-best}       & 36.3\% & 36.5\% & 27.2\%\\
\bottomrule
\end{tabular}
\caption{Head to head comparison between various systems. Models trained on the DailyDialog data outperform the filtered models, but there is no clear preference between base and best models.}
\label{tab:h2h}
\end{table*}

\subsection{Dialog Models}
\label{app:dialog_models}

We train Transformer conditional language models in \textsc{fairseq} using parameters from the \textsc{flores}\footnote{\url{https://github.com/facebookresearch/flores/tree/5696dd4ef07e29977d5690d2539513a4ef2fe7f0}} benchmark for low-resource machine translation \cite{guzman-etal-2019-flores}.

We use a  $5$-layer encoder and decoder, $512$ dimensional embeddings, and $2$ encoder and decoder attention heads. We regularize with $0.2$ label smoothing, and $0.4$ dropout. 
We optimize using Adam with a learning rate of $10^{-3}$. 
We train 100 epochs, and select the best checkpoint based on validation set perplexity. We generate with a beam size of $10$, and no length penalty. 

 \autoref{fig:nll_train} shows the train command.

\begin{figure*}[h]
\begin{verbatim}
python train.py \
 $DATADIR \
 --source-lang src \
 --target-lang tgt \
 --seed 10 \
 --save-dir $SAVEDIR \
 --patience 50 --criterion label_smoothed_cross_entropy \
 --label-smoothing 0.2 \
 --share-all-embeddings \
 --arch transformer  --encoder-layers 5 --decoder-layers 5 \
 --encoder-embed-dim 512 --decoder-embed-dim 512 \
 --encoder-ffn-embed-dim 2048 --decoder-ffn-embed-dim 2048 \
 --encoder-attention-heads 2 --decoder-attention-heads 2 \
 --encoder-normalize-before --decoder-normalize-before \
 --dropout 0.4 --attention-dropout 0.2 --relu-dropout 0.2 \
 --weight-decay 0.0001 \
 --optimizer adam --adam-betas '(0.9, 0.98)' --clip-norm 0 \
 --lr-scheduler inverse_sqrt --warmup-updates 4000 --warmup-init-lr 1e-7 \
 --lr 1e-3 --min-lr 1e-9 --no-epoch-checkpoints \
 --max-tokens 4000 \
 --max-epoch 100 --save-interval 10 --update-freq 4 \
 --log-format json --log-interval 100 
\end{verbatim}
\caption{Training command.}
\label{fig:nll_train}
\end{figure*}

We train and evaluate on the DailyDialog corpus \cite{li-etal-2017-dailydialog}, as released by ParlAI \cite{miller-etal-2017-parlai}.\footnote{\url{https://github.com/facebookresearch/ParlAI/tree/1e905fec8ef4876a07305f19c3bbae633e8b33af}} 

\subsection{Full Automatic Results}
\label{app:full}
\autoref{tab:hyperparam} shows the hyperparameters for each system.
\autoref{tab:multiref_word} and 
\autoref{tab:multiref_embed} show the evaluation against the multiple references for the word based and embedding based metrics. 
\autoref{tab:singleref_word} and 
\autoref{tab:singleref_embed} show the evaluation against the original single reference  for the word based and embedding based metrics. \autoref{tab:main_lexical_diversity} shows the lexical diversity, and \autoref{tab:ruq} shows the RUQ sores. 
\label{sec:appendix}

\begin{table*}[htb]
\addtolength{\tabcolsep}{-2pt}
\centering
\begin{tabular}{cc|ccccc}
\toprule
 Data& Params & bpe & lr       & dropout & otherdropout & labelsmooth \\ \midrule
\textsc{dd} & \textsc{base} & 4   & 0.001 & 0.4     & 0.2          & 0.2         \\
\textsc{dd} & \textsc{best} &4   & 0.001 & 0.0       & 0.1          & 0.4         \\
\textsc{ef} & \textsc{base} &4   & 0.001 & 0.4     & 0.2          & 0.2         \\
\textsc{ef} & \textsc{best} &2   & 0.001 & 0.0       & 0.1          & 0.2\\
\bottomrule
\end{tabular}
\caption{Hyperparameters for each of the four models we consider. }
\label{tab:hyperparam}
\end{table*}

\begin{table*}[htb]
\addtolength{\tabcolsep}{-3pt}
\begin{adjustbox}{max width=\linewidth}
\begin{tabular}{cc|cccc|cccc|cc}
\toprule
& & \multicolumn{4}{c|}{Average Max Sentence BLEU} & \multicolumn{4}{c|}{Corpus BLEU} & \multicolumn{2}{c}{}  \\ 
           Data& Params &BLEU1 & BLEU2 & BLEU3 & BLEU4 &BLEU1 & BLEU2 & BLEU3 & BLEU4& METEOR & ROUGE \\\midrule
\textsc{dd} & \textsc{base} &27.8          & 14.7          & 10.3          & 7.9           & 48.1          & 25.6          & 16.2          & 11.2          & 12.7          & 34.3          \\
\textsc{dd} & \textsc{best} &\textbf{33.9} & \textbf{21.9} & \textbf{17.7} & \textbf{15.3} & \textbf{53.9} & \textbf{36.1} & \textbf{28.9} & \textbf{25.1} & \textbf{17.8} & \textbf{39.7} \\
\textsc{ef} & \textsc{base} & 27.8          & 14.0          & 9.4           & 7.0           & 46.9          & 24.1          & 14.6          & 9.8           & 13.2          & 33.4          \\
\textsc{ef} & \textsc{best} & 31.7          & 19.1          & 14.9          & 12.7          & 51.0          & 32.8          & 25.5          & 21.8          & 16.9          & 37.2         \\\bottomrule
\end{tabular}
    \end{adjustbox}
\caption{Word-overlap based metrics on multiple references.}
\label{tab:multiref_word}
\end{table*}

\begin{table*}[htb]
\addtolength{\tabcolsep}{-2pt}
\centering
\begin{tabular}{cc|ccc|c}
\toprule
 && \multicolumn{3}{c|}{Cosine Similarity} & \\ 
 Data & Params              & SkipThought & Embed. Avg. &  VectorExtrema & GreedyMatching \\
               \midrule
\textsc{dd} & \textsc{base} &72.4          & 90.8            & 62.9          & 77.2          \\
\textsc{dd} & \textsc{best}&\textbf{73.8} & \textbf{92.2} & \textbf{65.4} & \textbf{79.3} \\
\textsc{ef} & \textsc{base}&71.9          & 91.2           & 62.2          & 77.0          \\
\textsc{ef} & \textsc{best}&72.8          & 91.6           & 62.7          & 77.9    \\     
\bottomrule
\end{tabular}
\caption{Embedding based metrics on multiple references.}
\label{tab:multiref_embed}
\end{table*}

\begin{table*}[]
\centering
\addtolength{\tabcolsep}{-3pt}
\begin{adjustbox}{max width=\linewidth}
\begin{tabular}{cc|cccc|cccc|cc}
\toprule
&& \multicolumn{4}{c|}{Average Max Sentence BLEU} & \multicolumn{4}{c|}{Corpus BLEU} & \multicolumn{2}{c}{}  \\ 
   Data & Params             &BLEU1 & BLEU2 & BLEU3 & BLEU4 &BLEU1 & BLEU2 & BLEU3 & BLEU4& METEOR & ROUGE \\
               \midrule
\textsc{dd} & \textsc{base} &15.3          & 7.6           & 5.6           & 4.5           & 12.9          & 6.3           & 4.1           & 3.0           & 6.7           & 20.6          \\
\textsc{dd} & \textsc{best} &\textbf{24.3} & \textbf{16.7} & \textbf{14.3} & \textbf{12.8} & \textbf{23.2} & \textbf{16.7} & \textbf{14.2} & \textbf{12.9} & \textbf{11.9} & \textbf{29.2} \\
 \textsc{ef} & \textsc{base} & 15.9          & 7.4           & 5.2           & 4.1           & 15.8          & 7.5           & 4.7           & 3.3           & 7.2           & 20.4          \\
 \textsc{ef} & \textsc{best} & 22.1          & 14.0          & 11.8          & 10.5          & 22.9          & 15.8          & 13.2          & 11.8          & 11.1          & 26.6   \\      

\bottomrule
\end{tabular}
    \end{adjustbox}

\caption{Word-overlap based metrics on the single reference test set.}
\label{tab:singleref_word}
\end{table*}

\begin{table*}[]
\addtolength{\tabcolsep}{-2pt}
\centering
\begin{tabular}{cc|ccc|c}
\toprule
& & \multicolumn{3}{c|}{Cosine Similarity} & \\ 
   Data     &   Params    & SkipThought & Embed. Avg. &  VectorExtrema & GreedyMatching \\
               \midrule
 \textsc{dd} & \textsc{base} & 65.3        & 86.3          & 50.6          & 71.3          \\
 \textsc{dd} & \textsc{best} &\textbf{68.2} &  \textbf{88.5} & \textbf{54.7} & \textbf{74.6} \\
 \textsc{ef} & \textsc{base} & 64.9       & 86.9          & 50.2          & 71.3          \\
 \textsc{ef} & \textsc{best} & 67.0        & 87.7          & 52.3          & 73.1\\       
\bottomrule
\end{tabular}
\caption{Embedding based metrics on the single reference test set.}
\label{tab:singleref_embed}
\end{table*}

\begin{table*}[]
\centering
\addtolength{\tabcolsep}{-2pt}
\begin{tabular}{cc|ccc}
\toprule
      Data     &   Params                                 & 1-grams & 2-grams & 3-grams \\\midrule
\textsc{dd}& \textsc{base} &2.4          & 10.3          & 18.8          \\
\textsc{dd}& \textsc{best}&3.5          & 18.0          &\textbf{35.5} \\
\textsc{ef}& \textsc{base} &2.3          & 10.7          & 20.1          \\
\textsc{ef}& \textsc{best}& \textbf{3.8} & \textbf{18.3} & 34.6     \\
\bottomrule

\end{tabular}

\caption{Type/Token ratios. }

\label{tab:main_lexical_diversity}
\end{table*}

\begin{table*}[]
\centering
\addtolength{\tabcolsep}{-2pt}
\begin{tabular}{cc|ccc}
\toprule
Data     &   Params                           & RUQ-train & RUQ-test  \\\midrule
\textsc{dd}& \textsc{base} &28.5          & 12.2          \\
\textsc{dd}& \textsc{best}&\textbf{95.3} & \textbf{35.7} \\
\textsc{ef}& \textsc{base}&37.9          & 15.5          \\
\textsc{ef}& \textsc{best}& 89.2          & 30.7         \\
\bottomrule

\end{tabular}
\caption{RUQ scores on the train and test data. }
\label{tab:ruq}
\end{table*}

\end{document}